\documentclass{midl} % Include author names

\usepackage{mwe} % to get dummy images
\usepackage{graphicx}
\usepackage{algorithm}
\usepackage{algorithmic}
\usepackage{array}
\usepackage{multirow}
\usepackage{amsmath}
\usepackage{amssymb}
\usepackage{pifont}
\usepackage{color}
\usepackage{CJKutf8}
\usepackage{caption}
\jmlrvolume{-- Under Review}
\jmlryear{2024}
\jmlrworkshop{Full Paper -- MIDL 2024 submission}
\editors{Under Review for MIDL 2024}

\title[Boundary-aware Contrastive Learning for Nuclei Segmentation]{Boundary-aware Contrastive Learning for Semi-supervised Nuclei Instance Segmentation}

\midlauthor{\Name{Ye Zhang\nametag{$^{1}$}}\Email{zhangye94@stu.hit.edu.cn} \\
		\Name{Ziyue Wang{$^{1}$}}\Email{200111326@stu.hit.edu.cn}\\
		\Name{Yifeng Wang{$^{2}$}} \Email{wangyifeng@stu.hit.edu.cn} \\
		\Name{Hao Bian{$^{3}$}} \Email{h2495067728@gmail.com}\\
		\Name{Linghan Cai{$^{1}$}} \Email{ceilinghans@gmail.cn}\\
		\Name{Hengrui Li{$^{4}$}} \Email{23B903028@stu.hit.edu.cn}\\
		\Name{Lingbo Zhang{$^{3}$}}\Email{zhang-lb23@mails.tsinghua.edu.cn}\\
		\Name{Yongbing Zhang{$^{1}$}} \Email{ybzhang@.hit.edu.cn}\\
		\addr $^{1}$ School of Computer Science and Technology, Harbin Institute of Technology, 518055, China.\\
		\addr $^{2}$ School of Science, Harbin Institute of Technology, 518055, China.\\
  	\addr $^{3}$ Tsinghua Shenzhen International Graduate School, Tsinghua University, 518071, China.\\
		\addr $^{4}$ Faculty of Computing, Harbin Institute of Technology, 150001, China.\\
}

\begin{document}

\maketitle

\begin{abstract}
Semi-supervised segmentation methods have demonstrated promising results in natural scenarios, providing a solution to reduce dependency on manual annotation. However, these methods face significant challenges when directly applied to pathological images due to the subtle color differences between nuclei and tissues, as well as the significant morphological variations among nuclei. Consequently, the generated pseudo-labels often contain much noise, especially at the nuclei boundaries. To address the above problem, this paper proposes a boundary-aware contrastive learning network to denoise the boundary noise in a semi-supervised nuclei segmentation task. The model has two key designs: a low-resolution denoising (LRD) module and a cross-RoI contrastive learning (CRC) module. The LRD improves the smoothness of the nuclei boundary by pseudo-labels denoising, and the CRC enhances the discrimination between foreground and background by boundary feature contrastive learning. We conduct extensive experiments to demonstrate the superiority of our proposed method over existing semi-supervised instance segmentation methods.
\end{abstract}


\section{Introduction}
Nuclei instance segmentation is essential in the quantitative analysis of pathological images. The characteristics of nuclei, including their size, morphology, and distribution, can provide valuable insights into the tumor immune microenvironment, thereby offering crucial support for cancer diagnosis, staging, and grading processes \cite{khened2021generalized,hollandi2022nucleus}. In recent years, deep learning techniques have made remarkable advancements in nuclei segmentation. DCAN \cite{chen2016dcan} adopts a dual-branch decoder architecture to predict semantics and contours simultaneously to enhance the instance distinguishing. HoverNet \cite{graham2019hover} incorporates distance and gradient constraints to split individual instances effectively. Similar methods such as Dist \cite{naylor2018segmentation}, CDNet \cite{he2021cdnet}, and CellPose \cite{stringer2021cellpose} are also proposed to address overlapping nuclei challenges. However, these supervised methods typically rely on pixel-level annotations, which are time-consuming and labor-intensive and need professional guidance, hindering the development of models. Therefore, developing a technique that can effectively address the dependency on manual annotation for nuclei instance segmentation is crucial.

A common approach to address the problem of scarce labeled data is semi-supervised learning \cite{reddy2018semi, van2020survey}. During the training process, abundant unlabeled and insufficient labeled data are used to train the network. The existing semi-supervised methods mainly leverage prior information to improve the pseudo-label quality. For example, ShapeProp \cite{zhou2020learning} combines the information from bounding boxes and partially annotated masks to improve the segmentation accuracy of target regions based on Mask R-CNN \cite{he2017mask}. PAIS \cite{hu2023pseudo} uses a dynamic alignment loss to address the misalignment problem between classification and segmentation results, and then a new threshold filtering method for pseudo-labels is proposed. PointWSSIS \cite{kim2023devil} balances false negative and false positive errors by utilizing point supervision prior information. However, due to the low color contrast differences between the nuclei and tissues, these methods still have defects in generating nuclear pseudo-labels, limiting the application of semi-supervised instance segmentation in pathological images.

Some methods use pseudo-label optimization strategies to enhance nuclei segmentation accuracy in semi-supervised scenarios. MMT-PSM \cite{zhou2020deep} integrates multiple data-augmented segmentation results to construct reliable predictions and enhance pseudo-labels' confidence. CDCL \cite{wu2022cross} uses feature contrastive learning to promote feature consistency between the teacher and student networks, thus improving the quality of pseudo-labels. PG-FANet \cite{jin2024inter} employs a pseudo-label guided module that aggregates multi-scale, multi-stage features to enhance segmentation performance. However, nuclei exhibit diversity in morphology and size, and in cases with limited annotations, it is challenging for the teacher network to capture the complete range of nuclei shape features. Consequently, the generated pseudo-labels often contain edge noise because existing pseudo-label optimization methods lack specific designs for denoising nuclei boundaries, which always leads to inaccurate nuclei boundary predictions.

In this paper, to address the issue of boundary noise in nuclei segmentation, we propose a coarse-to-fine \textbf{b}oundary-\textbf{a}ware contrastive learning network for \textbf{s}emi-supervised nuclei \textbf{s}egmentation (BASS\footnote[1]{Our code is availiable at \href{https://github.com/zhangye-zoe/BASS}{https://github.com/zhangye-zoe/BASS}.}). Firstly, we design a low-resolution denoising (LRD) segmentation head that promotes boundary smoothness. Additionally, within this segmentation head, we use a low-weight loss for the nuclei boundary region optimization, which reduces the impact of uncertain boundary prediction during training. Secondly, to minimize boundary noise further, we design a cross-RoI contrastive learning (CRC) module that finely partitions the internal, external, and boundary regions of nuclei, enhancing the discriminative capability of nuclei boundary features. To demonstrate the effectiveness of our proposed method, we conduct comparative experiments and ablation studies on two public datasets. The experimental results show that our proposed method outperforms existing semi-supervised methods, and the ablation studies demonstrate the effectiveness of the proposed modules.

\begin{figure*}[t!]
	\centering
	\includegraphics[width=5.5in]{"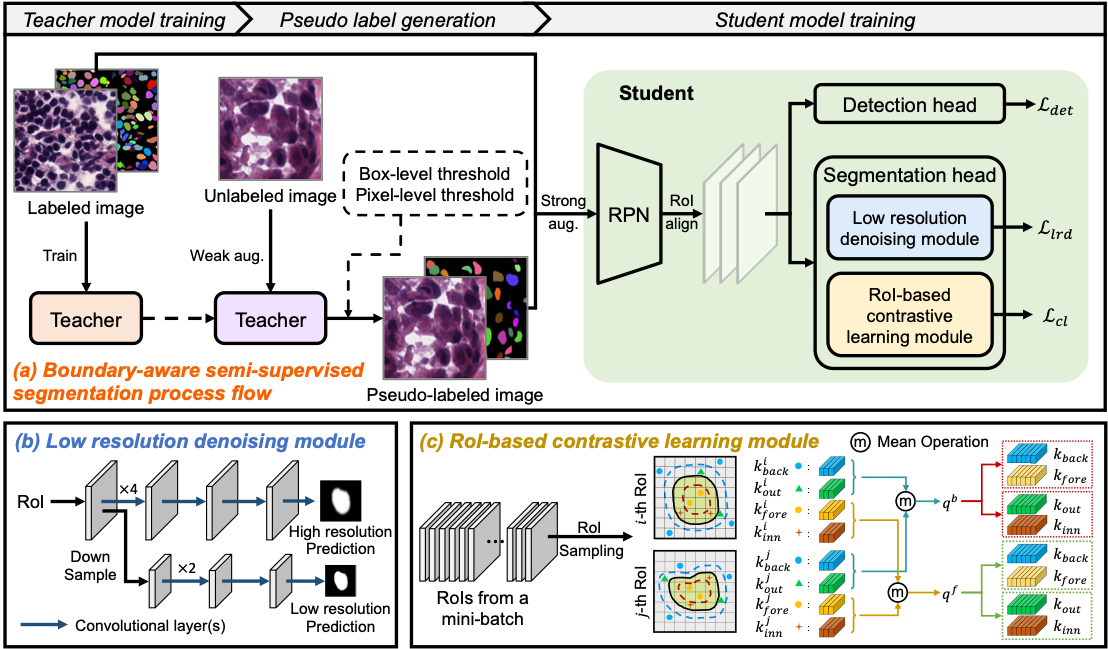"}
	\caption{The framework of our semi-supervised nuclei segmentation method. (a) The training flow of our BASS. First, the teacher model generates pseudo-labels, and then the student model is used to train the nuclei segmentation network. (b) and (c) is the proposed low-resolution denoising module and cross-RoI contrastive learning module.}
	\label{fig: flowchart}
 \vspace{-0.4cm}
\end{figure*}

\section{Methodology}
\subsection{Framework Overview}

To address the boundary noise problem of nuclei segmentation under a semi-supervised scenario, we propose a coarse-to-fine boundary-aware denoising model, as shown in Fig. \ref{fig: flowchart}. 
Our whole training process can be divided into three stages. 
First, the labeled data $D_L = \{ {(x_i, y_i)}\}_{i=1}^{N}$ is used to train a teacher model. In this step, the optimized loss for the teacher network is defined as follows:
\begin{equation}
	Loss^{t}= L_{seg}^{t} + L_{det}^{t},
\end{equation}
where $L_{seg}^{t}$ is the loss of the segmentation head, and $L_{det}^{t}$ is the loss of the detection head, which consists of the classification loss and regression loss. Then, the trained teacher network is employed to generate pseudo-labels $y_j^p$ for input $x_j$. To reduce the uncertainty of pseudo-labels, we employ box and pixel threshold filtering to generate high-confidence pseudo-labels.
Finally, we combine the labeled data $D_L$ and the generated pseudo-labeled data $D_U = \{(x_j, y_j^p)\}_{j=1}^M$ to train the student network. 

To avoid the impact of noisy pseudo-labels in the student network, we introduce coarse-to-fine boundary denoising methods as shown in the green box of Fig. \ref{fig: flowchart}. The methods consist of a coarse low-resolution denoising (LRD) module and a fine-denoising cross-RoI contrastive learning (CRC) module. The LRD employs low-resolution pseudo-labels as supervision information to promote the smoothness of nuclei contours. Meanwhile, the CRC utilizes boundary-aware contrastive learning to enhance the discriminative capability of contour features. In the training process, the overall loss is designed as follows:
\begin{equation}
	Loss^s = L_{det}^{s} + \omega_1 L_{nmh} + \omega_2 L_{lrd} + \omega_3 L_{cl},
\end{equation}
where $L_{nmh}$ represents the naive high resolution segmentation loss, $L_{lrd}$ represents the low-resolution segmentation loss, and $L_{cl}$ represents the contrastive learning loss. $\omega_1$, $\omega_2$ and $\omega_3$ are used to balance the loss of weight. In the paper, $\omega_1$, $\omega_2$ and $\omega_3$ are set to 1.

\subsection{Low-resolution Denoising Module}

Due to the issue of boundary noise in pseudo-labels, the student network is simply affected during the training process, and this influence gradually accumulates and impacts the final segmentation results. To mitigate the impact of boundary noise, BASS employs a coarse-denoising strategy. Firstly, we apply box and pixel thresholds to filter the predicted results, generating pseudo-labels with high confidence. The generated pseudo-labels are used to supervise training in the subsequent process. Next, we design a low-resolution denoising module, which utilizes the low-resolution pseudo-labels as supervision for model training, as shown in Fig. \ref{fig: flowchart}(b). 

In the naive Mask R-CNN \cite{he2017mask}, the RoI head outputs a $14 \times 14$ feature map containing boundary noise. In the subsequent convolution process, Mask R-CNN increases the feature map to capture semantic information in a larger size, but the boundary noise is also amplified. In contrast, BASS directly performs segmentation in the $14 \times 14$ feature map to avoid the influences of enlarged noise in convolution. This approach effectively smooths the boundaries and initially reduces the noise in nuclei boundaries. Furthermore, to minimize the impact of boundary uncertainty on segmentation, we apply a weighted loss to the low-resolution segmentation head. Specifically, pixels in the boundary region are assigned a lower weight, and other areas are set to a high weight.

According to previous studies \cite{wang2022noisy}, although low-resolution images can reduce the boundary noise, they lose some detailed information. To preserve the details, we parallel the original segmentation head and low-resolution prediction head to perform the segmentation task simultaneously, as shown in Fig. \ref{fig: flowchart}(b). In this manner, the output mask head decreases the influence of the original feature noise and keeps more details.

\subsection{Cross-RoI Contrastive Learning}

In the subsection, we propose an elaborate denoising method named cross-RoI contrastive learning. It leverages labeled data to train a boundary feature extraction module, and then the module is applied to learn the embedding of unlabeled data, which can mitigate the impact of boundary noise caused by pseudo-labels and enhance the feature discrimination ability of foreground and background. Our proposed CRC is shown in Fig. \ref{fig: flowchart} (c). 

First, the input image $x$ is fed into the network for feature extraction and alignment, resulting in two aligned RoI features $f_i$ and $f_j$. To address the challenges of differentiating nuclei from tissues due to staining variations and the issue of boundary noise, we employ pixel contrastive learning to optimize the foreground and background features. However, object boundaries typically correspond to hard-to-classify samples, and their embeddings are highly unstable. To avoid the impact of features from difficult samples on the representation of easy-to-classify samples, we employ the approach of region-based contrastive learning. In detail, we split the feature map $f_i$ and $f_j$ into four regions according to inner and outer contours. The inner contour is plotted with the red dotted line, and the outer contour is plotted with the blue dotted line shown in Fig. \ref{fig: flowchart} (c). 
Based on the inner and outer contours, the sampled inner contour pixel set $P_{inn}$ and outer contour pixel set $P_{out}$ are described as:
\begin{equation}
\begin{split}
    P_{inn} = \{p_i| \; {p_i \in P_{fore} \ and \  \Vert p_i, c_i \Vert}_{2}^{2} \leq d\}, \\
    P_{out} = \{p_i| \; {p_i \in P_{back} \ and \  \Vert p_i, c_i \Vert}_{2}^{2} \leq d\} , 
\end{split}
\end{equation}
where $P_{fore}$ represents the foreground pixel set, $P_{back}$ represents background pixel set, and $c_i$ represents the contour pixel closest to pixel $p_i$. At the same time, we also sample the foreground and background pixels, which can be expressed as the following equations:
\begin{equation}
\begin{split}
    P_{fore-inn} = \{ p_i | \; p_i \in P_{fore} \ and \ p_i \not\in P_{inn} \},\\
    P_{back-out} = \{ p_i | \; p_i \in P_{back} \ and \ p_i \not\in P_{out} \},
\end{split}	
\end{equation}
where $P_{fore-inn}$ represents the set of pixels obtained by excluding  $P_{inn}$ from $P_{fore}$ and  $P_{back-out}$ represents the set of pixels obtained by excluding $P_{out}$ from $P_{back}$.

Next, we sample pixel features from the sets $P_{inn}$, $P_{out}$, $P_{fore-inn}$ and $P_{back-out}$. The sampling ratio is set to $\alpha$. For feature $f_i$, the sampled features are denoted as $k_{back}^i$, $k_{out}^i$, $k_{fore}^i$ and $k_{inn}^i$ respectively. Similarly,  for feature $f_j$, the sampled features are denoted as $k_{back}^j$, $k_{out}^j$, $k_{fore}^j$ and $k_{inn}^j$ respectively.

Then, we calculate the query features of background and foreground across RoIs through: 
\begin{equation}
    q^b = M(k_{back}^i, k_{out}^i, k_{back}^j, k_{out}^j), \quad q^f = M(k_{fore}^i, k_{inn}^i, k_{fore}^j, k_{inn}^j),
\end{equation}
where $M$ represents the averaged operation of vectors.

Finally, to narrow the same category feature distance and expand the feature distance between foreground and background. We calculate four pairs of contrastive losses as follows:
\begin{equation}
    L_{cl}^{s} = CL(q^b, k_{back}, k_{fore}) + CL(q^b, k_{out}, k_{inn})+ CL(q^f, k_{fore}, k_{back}) + CL(q^f, k_{inn}, k_{out}), 
\end{equation}
where $CL$ represents contrastive learning loss. $k_{back}$ represents the concatenation of $k_{back}^{i}$ and $k_{back}^{j}$. The calculation of $CL$ is described below:
\begin{equation}
    CL(q^+, k^+, k^-) = -log{\frac{e^{cos(q^+, k^+)/\tau}}{e^{cos(q^+,k^+)/\tau}+\sum_{i=1}^{N}e^{cos(q^+, k^-_{i})/\tau}}},
\end{equation}
where $q^+$ and $k^+$ represent a pair of positive instances, $k^-$ represents a negative instance, and $\tau$ is the temperature hyper-parameter.

Different from the previous methods, PC$_2$Seg \cite{zhong2021pixel} extracts positive instance pair of contrastive learning from a single perspective. However, our proposed CRC performs contrastive learning cross-RoI, which enhances the feature generality.

\section{Experiments}
\subsection{Datasets}
Our method is evaluated on the Cryosectioned and H\&E-stained Nuclei Segmentation (CryoNuSeg) dataset \cite{mahbod2021cryonuseg} and the Digestive-System Pathological Segmentation (DigestPath) dataset \cite{da2022digestpath}. CryoNuSeg contains 30 images from 10 organs, each with a size of $512\times512$. We randomly select 18 images for training, 6 for validation, and 6 for testing. DigestPath contains 69 images of the digestive system, each with a size of approximately 1500$\times$1200. We randomly select 41 images for training, 14 for validation, and 14 for testing.

\vspace{-0.3cm}
\subsection{Implementation Details and Evaluation Metrics}
Following the previous method \cite{graham2019hover}, we crop all the images to patches of 256 × 256 pixels with an overlap of 128 pixels for data preprocessing. All experiments are carried out with an RTX 3090 GPU. SGD is used as the optimizer. The learning rate, momentum, and weight decay are set to 0.02, 0.9, and 0.001, respectively. Besides, we evaluate the segmentation performance in terms of Dice \cite{vu2019methods}, aggregated Jaccard index (AJI) \cite{kumar2017dataset}, and panoptic quality (PQ) \cite{kirillov2019panoptic}. 

\begin{table*}[ht]\footnotesize
% \footnotesize
\centering
	\renewcommand{\arraystretch}{1.05}
	\caption{Performance comparisons on CryoNuSeg and DigestPath Datasets. The best performance is highlighted in \textbf{bold}, and the second-best is \underline{underlined}}
	% \resizebox{\linewidth}{90pt}{
		\begin{tabular}{c|c|ccc|ccc}
			\hline
			  \textbf{Amount} &\multirow{2}{*}{\textbf{Methods}} & \multicolumn{3}{c|}{\textbf{CryoNuSeg}} & \multicolumn{3}{c}{\textbf{DigestPath}} \\
			\cline{3-8}
			\textbf{of labels}& & \textbf{Dice} &\textbf{AJI} & \textbf{PQ}  & \textbf{Dice} &\textbf{AJI} & \textbf{PQ} \\
			\hline
			& MMT-PSM \cite{zhou2020deep} & 54.83 & 30.17 & 29.81 & 55.68 & 32.34  & 35.94 \\
			& PointWSSIS \cite{kim2023devil} & \underline{58.66}& 35.41 & 33.61 & \underline{59.90}& \underline{40.06} & \underline{44.10} \\
			\textbf{1/8} & ShapeProp \cite{zhou2020learning} & 57.42 &\underline{35.53} & \underline{33.68} & 58.18& 39.94 & 43.49 \\
			& NoisyBoundary \cite{wang2022noisy} & 55.14 & 29.57 & 30.96 & 58.34 & 36.75 & 37.94 \\
			& BASS (Ours) & \textbf{59.26}&\textbf{36.32} &\textbf{35.09} & \textbf{61.00}& \textbf{41.33} & \textbf{45.07} \\
			\cline{1-8}
			& MMT-PSM \cite{zhou2020deep} & 67.24 & 37.60 & 34.67 & 58.23 & 37.64 & 41.93  \\
			& PointWSSIS \cite{kim2023devil} & \textbf{75.01} & 47.12 & \underline{49.83} & \textbf{64.93}& 43.16 & 47.86\\
			  \textbf{1/4} & ShapeProp \cite{zhou2020learning} & 73.37 &\underline{48.70} & 48.72 & 63.31& \underline{43.35} & \underline{48.44} \\
			& NoisyBoundary \cite{wang2022noisy} & 69.34 &38.85 & 35.91& 61.15 & 40.77& 45.74 \\
			& BASS (Ours) &\underline{74.79} &\textbf{48.96} & \textbf{50.36} & \underline{63.41} & \textbf{44.72} & \textbf{49.14} \\
			\cline{1-8}
			& MMT-PSM \cite{zhou2020deep} & 72.85 & 45.06 & 44.47 &  59.11&39.97 & 45.58\\
			& PointWSSIS \cite{kim2023devil} &  \underline{74.67}& \underline{49.91} & \underline{49.29} &  63.87&\underline{45.45} & 51.64 \\
			\textbf{1/2} & ShapeProp \cite{zhou2020learning} & 74.40 & 48.24 & 47.55 & \underline{64.15} & 45.02 & \underline{52.93}\\
			& NoisyBoundary \cite{wang2022noisy} & 73.71 &46.58 & 46.13 &  61.35& 44.41& 50.65  \\
			& BASS (Ours) & \textbf{76.76}& \textbf{51.09} & \textbf{49.66} &  \textbf{65.72} & \textbf{46.14} & \textbf{53.96}  \\
			\hline
    		\end{tabular}
		% \end{center}
	\label{tab1}
\end{table*}
\vspace{-0.4cm}
\subsection{Comparison with the State of the Art Methods}
We employ a Mask R-CNN as the baseline of the semi-supervised segmentation methods. We compare our proposed BASS against several state-of-the-art methods, including MMT-PSM \cite{zhou2020deep}, PointWSSIS \cite{kim2023devil}, ShapeProp \cite{zhou2020learning} and NoisyBoundary \cite{wang2022noisy}. We trained the models using 1/8, 1/4, and 1/2 of the labeled data on ResNet-50. A more detailed data split is presented in the appendix.

\begin{figure*}
    \centering
    \includegraphics[width=5.2in]{"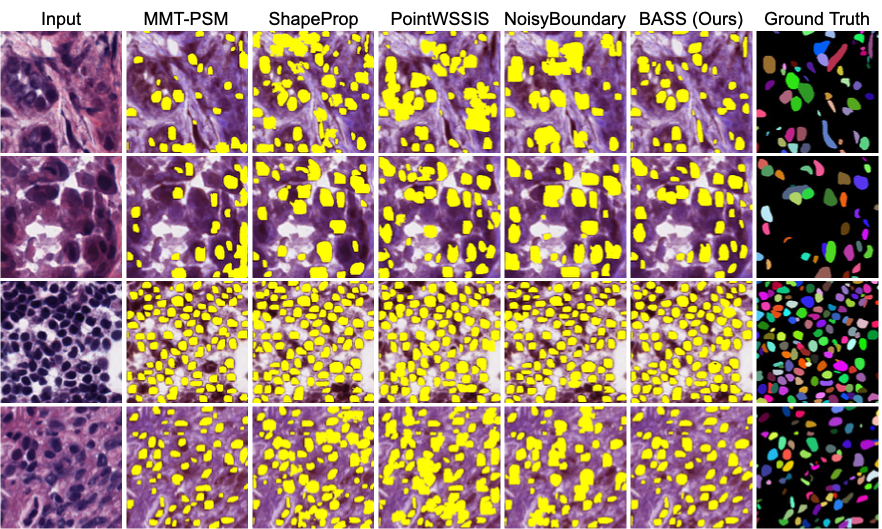"}
    \caption{The semi-supervised instance segmentation visualization comparisons.}
    \label{fig: visulization}
    \vspace{-0.3cm}
\end{figure*}

Quantitative comparison results on CryoNuSeg and DigestPath are displayed in Table \ref{tab1}, which shows that our method achieves the optimal performance at all three annotation ratios. Even with only 1/8 of the annotations, our BASS improves the PQ metric by approximately 1\%, demonstrating our method's superiority. Fig. \ref{fig: visulization} displays the visual comparison results among existing semi-supervised segmentation models. From the figure, we can find that MMT-PSM and NoisyBoundary mistakenly identify nuclei as tissue due to the lack of semantic discrimination between nuclei and tissues. Although ShapeProp and PointWSSIS employ weak labels to enhance the location ability of nuclei, they still have issues with nucleus shape errors. In contrast, BASS can segment nuclei precisely. These results further demonstrate the effectiveness of our boundary-aware denoising method in the semi-supervised nuclei segmentation task.

\vspace{-0.4cm}
\subsection{Ablation Studies}
\vspace{-0.2cm}
\noindent\textbf{Ablation Studies for Segmentation Head.} 
In the student network, we employ three prediction heads, namely, the naive mask head (NMH), low-resolution denoising mask head (LRD), and cross-RoI contrastive learning mask head (CRC), to jointly supervise the segmentation predictions. To evaluate the effectiveness of these heads, we conducted a series of ablation experiments to assess the impact of different designs.
Specifically, we compared four designs: NMH, LRD, NMH+LRD, and NMH+LRD+CRC. The experimental results are listed in Table \ref{tab2}. The results demonstrate that NMH+LRD+CRC outperforms the other methods, indicating that incorporating multiple segmentation constraints is effective. Secondly, we observed a significant drop in performance when using NMH or LRD alone. However, the performance improves when combining NMH and LRD, suggesting that NMH and LRD can synergistically enhance segmentation performance. NMH provides more detailed information, while LRD reduces noise in nuclei boundaries. Lastly, comparing NMH+LRD+CRC with NMH+LRD, we found a significant improvement in the former, indicating the effectiveness of the CRC module. These findings highlight the collaborative impact of NMH and LRD in improving segmentation performance, with additional benefits arising from integrating the CRC module.

\noindent\textbf{Ablation Studies for Sampling Ratio $\alpha$.} 
We further conduct ablation experiments on the sampling ratio in the CRC. We set four sampling ratios of 0.1, 0.3, 0.5, and 0.7. The experimental results are shown in Table \ref{tab3}. The table shows that as the sampling ratio increases, the performance gradually improves, indicating that the sampling ratio indeed influences the performance. When the sampling ratio is large, the model obtains more sampled pixels, resulting in better contrastive learning performance. However, as the sampling ratio increases, the computational cost of the model also increases. Therefore, we select 0.7 as the final sampling ratio, which achieves the best balance between model performance and computational cost.

\begin{table}[ht]\footnotesize
\centering
\vspace{-0.2cm}
\caption{The segmentation head ablation experiments on CryoNuSeg and DigestPath datasets.}
    \begin{tabular}{ccc|ccc|ccc}
        \hline
        \multirow{2}{*}{\textbf{NMH}} & \multirow{2}{*}{\textbf{LRD}} & \multirow{2}{*}{\textbf{CRC}} & \multicolumn{3}{c}{\textbf{CryoNuSeg}} & \multicolumn{3}{c}{\textbf{DigestPath}} \\
        \cline{4-9}
        & & & \textbf{Dice} &\textbf{AJI} & \textbf{PQ} & \textbf{Dice} &\textbf{AJI} & \textbf{PQ}\\
        \hline
        \ding{51} & & & 65.49 & 40.23  & 36.94 & 60.80 & 44.73 & 48.54  \\
        & \ding{51} &  & 64.37 & 39.24 & 39.91 & 61.22 & 43.09 & 47.80 \\
        \ding{51} & \ding{51} & & \underline{72.17}& \underline{46.44} & \underline{45.72} & \underline{63.19}& \underline{45.86} & \underline{51.22} \\
        \ding{51} & \ding{51} & \ding{51} & \textbf{76.76}& \textbf{51.09} & \textbf{49.66} & \textbf{65.72} & \textbf{46.14} & \textbf{53.96} \\
        \hline
\end{tabular}\label{tab2}
\end{table}

\noindent\textbf{Ablation Studies for Distance $d$.}
In this part, we investigate the effect of distance $d$, which represents the distance from the inner (outer) contour to the accurate nuclei contour. Expressly, we set $d$ to 0, 2, 4, and 6. It is worth noting that when $d=0$, we do not sample between the actual and inner (outer) contour. The experimental results are shown in Table \ref{tab4}. From the table, we can see when $d=4$, the model performs best. However, the performance drops as $d$ decreases. This is because when reducing the sampling distance, the boundary information obtained by the model also decreases. On the contrary, when $d$ increases to 6, the sampling area of the boundary becomes more extensive, leading to a mixture of boundary and non-boundary features, ultimately decreasing performance. Therefore, a reasonable sampling distance is critical to model performance.

\vspace{15pt}
\begin{minipage}[c]{0.45\textwidth}
\centering
\tiny
\vspace{-0.2cm}
\captionof{table}{The sampling ratio ablation experiments on CryoNuSeg dataset.}
\resizebox{\linewidth}{34pt}{
\begin{tabular}{c|ccc}
\hline
\textbf{Ratio}& \textbf{Dice} &\textbf{AJI} & \textbf{PQ} \\
\hline
    $\alpha=0.1$ & 71.87 & 47.05 & 45.80 \\
    $\alpha=0.3$ & 73.05 & 47.70 & 46.26 \\
    $\alpha=0.5$ & \underline{76.56}&\underline{50.64} & \underline{48.94} \\
    $\alpha=0.7$ & \textbf{76.76}&\textbf{51.09} &\textbf{49.66}\\
\hline
    \end{tabular}}\label{tab3}
\vspace{-0.3cm}
\end{minipage}
\hspace{15pt}
\begin{minipage}[c]{0.45\textwidth}
\centering
\tiny
\vspace{-0.2cm}
\captionof{table}{The distance comparison experiments on CryoNuSeg dataset.}
\resizebox{\linewidth}{34pt}{
\begin{tabular}{c|ccc}
\hline
\textbf{Distance}& \textbf{Dice} &\textbf{AJI} & \textbf{PQ} \\
\hline
    $d = 0$ & 74.89 & 48.72 & 42.42 \\
    $d = 2$ & \underline{76.53} & \underline{50.23} & \underline{47.71} \\
    $d = 4$ & \textbf{76.76}& \textbf{51.09} & \textbf{49.66} \\
    $d = 6$ & 76.08&49.31 &47.22 \\
\hline
\end{tabular}}\label{tab4}
\vspace{-0.3cm}
\end{minipage}

\section{Conclusions}
This paper proposes a boundary-aware contrastive learning model for semi-supervised nuclei segmentation based on the teacher-student framework. In the student network, a low-resolution denoising module and a cross-RoI contrastive learning module are proposed to ease the contour noises of nuclei from coarse and fine aspects. Extensive comparison experiments on two publicity datasets show that the proposed method is superior to the existing semi-supervised instance segmentation methods.

% \bibstyle:{plain}
\bibliography{MIDL2024-DBCNet/midl}
\newpage
\section{Supplement Materials}
\subsection{Data Split}

In the main body, we used 1/8, 1/4 and 1/2 labeled data to conduct experiments on CryoNuSeg \cite{mahbod2021cryonuseg} and DigestPath \cite{da2022digestpath} datasets respectively. 

In this section, we provide the data split details as shown in Table \ref{tab1}. 
First, these two datasets are divided into the training set, validation set and testing set according to the proportion of 6:2:2. Then, we re-divide the training set into labeled and unlabeled data sets according to 1/8, 1/4 and 1/2. In the whole training process, we keep the validation and testing sets unchanged.

\begin{table}[thp]\footnotesize
\begin{center}
\caption{The data split on CryoNuSeg and DigestPath datasets.}
        \begin{tabular}{cc|cc|cc}
            \hline
            \multirow{2}{*}{\textbf{Dataset}} & \multirow{2}{*}{\textbf{Ratio}} & \multicolumn{2}{c|}{\textbf{Training}} & \multirow{2}{*}{\textbf{Validation}} & \multirow{2}{*}{\textbf{Testing}} \\
            \cline{3-4}
            & & \textbf{Labeled} & \textbf{Unlabeled} & &  \\
            \hline
            & \textbf{1/8} & 20 &142 & 54 & 54  \\
            \textbf{CryoNuSeg} & \textbf{1/4} & 40 & 122 & 54 & 54 \\
            & \textbf{1/2} & 81 & 81 & 54& 54 \\
            \hline
            & \textbf{1/8} & 631 & 2653 & 835 & 994  \\
            \textbf{DigestPath} & \textbf{1/4} & 930 &2354 & 835 & 994 \\
            & \textbf{1/2} & 1740 & 1554 & 835 & 994 \\
            \hline
        \end{tabular}
    \end{center}
    \label{tab5}
\end{table}

\subsection{Segmentation head visualization experiments}

\begin{figure}[tph]
	\centering
	\includegraphics[width=5.0in]{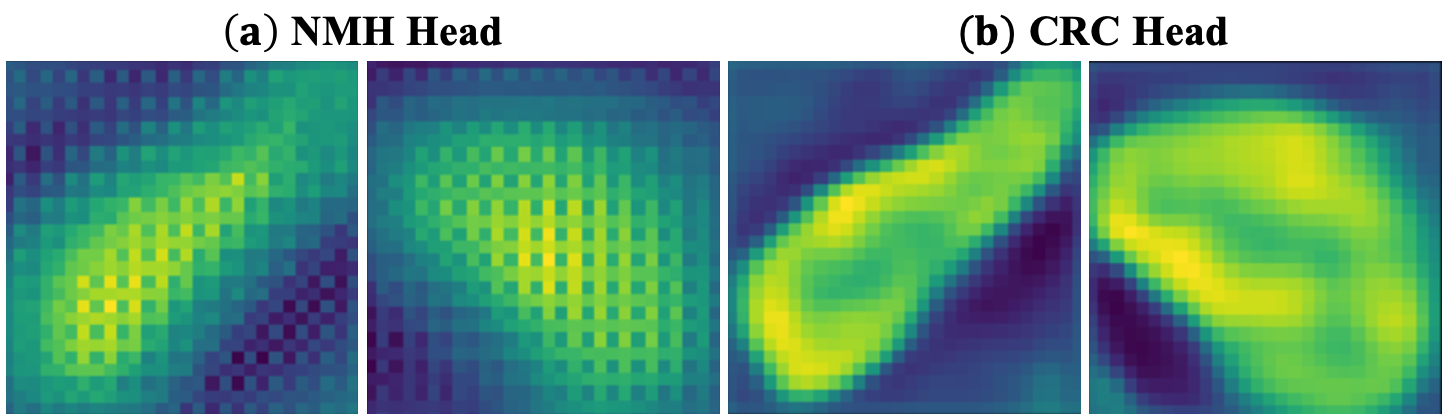}
	\caption{The feature map of prediction. (a) represents naive mask head (NMH); (b) represents cross-RoI contrastive learning head (CRC).}
	\label{mmc visual}
\end{figure}

In this subsection, we provide visual experiments of the naive segmentation head (NMH) and the cross-RoI contrastive learning head (CRC), as shown in Fig. \ref{mmc visual}. By comparing with NMH, it can be observed that the CRC module, utilizing contrastive learning, reduces edge noise and provides clearer boundaries.

\subsection{Comparison experiments}
Due to the limited space in the main text, we have provided additional experimental results using ResNet-101 as the backbone in Table \ref{tab6}. From the table, it can be observed that our method outperforms other approaches and achieves the best performance.

\begin{table*}[ht]\footnotesize
\centering
	\renewcommand{\arraystretch}{1.1}
	\caption{Performance comparisons on CryoNuSeg and DigestPath Datasets.}
        \begin{tabular}{c|c|ccc|ccc}
        \hline
        \textbf{Amount} &\multirow{2}{*}{\textbf{Method}} & \multicolumn{3}{c|}{\textbf{CryoNuSeg}} & \multicolumn{3}{c}{\textbf{DigestPath}} \\
        \cline{3-8}
        \textbf{of Labels}& & \textbf{Dice} &\textbf{AJI} & \textbf{PQ}  & \textbf{Dice} &\textbf{AJI} & \textbf{PQ} \\
        \hline
        & MMT-PSM \cite{zhou2020deep} &  56.66 & 33.96 & 30.81 &57.37 &36.66 & 39.55 \\
        & PointWSSIS \cite{kim2023devil}  &59.24  & \underline{36.13} &35.59 & \underline{62.91} & \textbf{43.19} & 46.36 \\
        \textbf{1/8} & ShapeProp \cite{zhou2020learning} & \textbf{60.07}&35.44 & \underline{35.90} & 61.22 & 41.69& \underline{46.57}\\
        & NoisyBoundary \cite{wang2022noisy} & 57.12& 32.78 & 34.66 & 59.11 & 38.31 & 40.46  \\
        & BASS(Ours)  & \underline{60.03}& \textbf{36.42} & \textbf{36.06} & \textbf{63.31} & \underline{42.17} & \textbf{47.25}  \\
        \cline{1-8}
        & MMT-PSM \cite{zhou2020deep} &71.27 &38.98 &36.28 &60.73 & 72.22& 44.23  \\
        & PointWSSIS \cite{kim2023devil} &\underline{76.39} &48.67 &\underline{50.96} & \underline{64.71} & 43.24 & \underline{48.48}  \\
          \textbf{1/4} & ShapeProp \cite{zhou2020learning} & 75.26&\underline{48.99} & 49.79 & 64.14& \underline{44.12} &48.00 \\
        & NoisyBoundary \cite{wang2022noisy} & 71.78 & 40.05 & 38.31 & 62.84& 43.71 & 46.56 \\
        & BASS(Ours) & \textbf{76.81} & \textbf{50.72} & \textbf{51.20} & \textbf{65.12}& \textbf{45.87} & \textbf{50.91} \\
        \cline{1-8}
        & MMT-PSM \cite{zhou2020deep} & 71.58 &47.21 &48.19& 62.79&43.53 & 50.04  \\
        & PointWSSIS \cite{kim2023devil} & \underline{76.89} &\underline{49.31} &\underline{52.42} & \underline{65.35} & \textbf{48.26} & \underline{52.56}   \\
        \textbf{1/2} & ShapeProp \cite{zhou2020learning} &76.40  &48.11 &51.21& 64.26& 46.04 & 51.64 \\
        & NoisyBoundary \cite{wang2022noisy} & 74.02 & 48.13 & 49.34 & 64.41&47.23 & 52.18 \\
        & BASS(Ours) & \textbf{77.14} & \textbf{51.06} & \textbf{54.39}  &\textbf{66.89} & \underline{48.10} & \textbf{55.59} \\
        \cline{1-8}
        \end{tabular}
	\label{tab6}
\end{table*}

\end{document}